\documentclass[conference]{IEEEtran}
\IEEEoverridecommandlockouts
\usepackage{cite}
\usepackage{amsmath,amssymb,amsfonts}
\usepackage{algorithm}
\usepackage{multirow}
\usepackage{algpseudocode}
\usepackage{graphicx}
\usepackage{subfigure}
\usepackage{textcomp}
\usepackage{xcolor}
\usepackage{caption}
\usepackage[normalem]{ulem}
\useunder{\uline}{\ul}{}

\def\BibTeX{{\rm B\kern-.05em{\sc i\kern-.025em b}\kern-.08em
    T\kern-.1667em\lower.7ex\hbox{E}\kern-.125emX}}

\title{Multi-omics Sampling-based Graph Transformer for Synthetic Lethality Prediction}

\vspace{-4mm}
\author{\IEEEauthorblockN{Xusheng~Zhao\textsuperscript{1,2},
Hao~Liu\textsuperscript{1,2},
Qiong~Dai\textsuperscript{1*}, 
Hao~Peng\textsuperscript{3},
Xu~Bai\textsuperscript{1},
Huailiang~Peng\textsuperscript{1}}
\IEEEauthorblockA{\textsuperscript{1}Institute of Information Engineering, Chinese Academy of Sciences, Beijing, China\\
\textsuperscript{2}School of Cyber Security, University of Chinese Academy of Sciences, Beijing, China\\
\textsuperscript{3}School of Cyber Science and Technology, Beihang University, Beijing, China\\
\textsuperscript{*}Corresponding author\\
\{zhaoxusheng, liuhao1998, daiqiong\}@iie.ac.cn, penghao@buaa.edu.cn, \{baixu, penghuailiang\}@iie.ac.cn
}}

\begin{document}

\maketitle
\begin{abstract}
Synthetic lethality (SL) prediction is used to identify if the co-mutation of two genes results in cell death.
The prevalent strategy is to abstract SL prediction as an edge classification task on gene nodes within SL data and achieve it through graph neural networks (GNNs).
However, GNNs suffer from limitations in their message passing mechanisms, including over-smoothing and over-squashing issues.
Moreover, harnessing the information of non-SL gene relationships within large-scale multi-omics data to facilitate SL prediction poses a non-trivial challenge.
To tackle these issues, we propose a new multi-omics sampling-based graph transformer for SL prediction (MSGT-SL).
Concretely, we introduce a shallow multi-view GNN to acquire local structural patterns from both SL and multi-omics data.
Further, we input gene features that encode multi-view information into the standard self-attention to capture long-range dependencies.
Notably, starting with batch genes from SL data, we adopt parallel random walk sampling across multiple omics gene graphs encompassing them.
Such sampling effectively and modestly incorporates genes from omics in a structure-aware manner before using self-attention.
We showcase the effectiveness of MSGT-SL on real-world SL tasks, demonstrating the empirical benefits gained from the graph transformer and multi-omics data.
\end{abstract}


\section{Introduction}\label{Sec_1}
Synthetic lethality (SL) denotes a class of gene relationships in which two genes mutated or lost together result in cell death, while a mutation in either gene alone does not.
Such properties make SL shine in targeted cancer therapy, whose key treatment is to interfere with the normal partner gene of a defective gene in cancer cells and kill them.
For instance, as a PARP inhibitor, olaparib targets breast cancer in patients with BRCA mutations by exploiting the SL relationship between PARP and BRCA1/2 genes \cite{lord2017parp}.
However, because a large number of gene pairs with potential SL relationships are unidentified \cite{wang2022computational}, it is imperative to develop approaches for SL prediction.

In the early stages of SL research, several prediction avenues were pursued to identify if mutations in pairs of genes together lead to cell death \cite{beijersbergen2017synthetic}.
As an early and widely applied approach, genetic screens \cite{hartman2001principles,kelley2005systematic,brough2011functional} manipulate gene expression or fabricate mutations to uncover SL relationships among specific genes.
In contrast to inefficient screening, the high-throughput approach \cite{tong200716,brough2011searching} applies automated equipment and large-scale sampling techniques to evaluate numerous gene pairs.
The advancements in functional genomics (e.g., RNAi \cite{luo2009genome} and CRISPR-Cas9 gene editing \cite{du2017genetic,wang2017gene}) provide a more comprehensive understanding of SL.
These wet-lab methods achieve biological validation but often require advanced techniques, specialized equipment, and significant investments of time and resources.

To reduce costs, computational dry-lab methods have gained extensive attention for their efficiency in processing large-scale gene data \cite{wang2022computational}.
In particular, they typically leverage rule-based statistical inference, gene graph analysis, and machine learning models \cite{tang2022synthetic}.
Based on the definition of SL, statistical inference \cite{jerby2014predicting,lee2018harnessing,liany2020aster,sinha2017systematic} utilizes statistical tests to analyze gene data and infer lethal pairs.
Graph-based analysis \cite{alvarez2016functional,hu2019optimal} regards top-ranked gene pairs as cell killers through protein-protein and metabolic networks, etc. 
In addition, machine learning models trained on clear SL pairs have emerged in SL detection, including support vector machine \cite{cho2016compact,gao2013integrative}, random forest \cite{li2019identification,das2019discoversl,benstead2019predicting}, and matrix factorization \cite{liany2020predicting,huang2019predicting,liu2019sl}.
While these shallow methods can offer clearer reasoning and explanations, they are often restricted by complex feature engineering and linear assumptions.

\begin{figure*}[t]
	\centering
	\includegraphics[width=\textwidth]{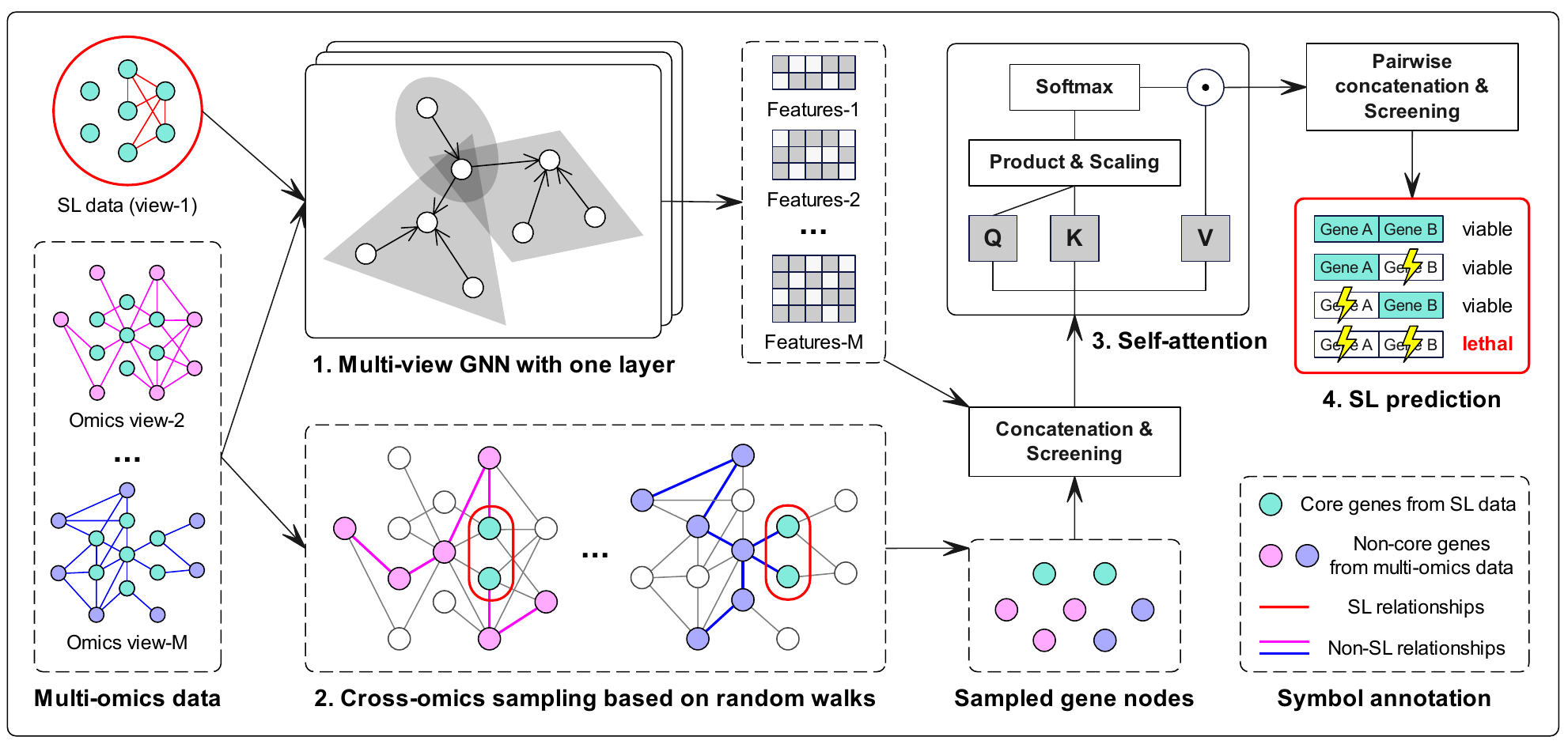}
    \vspace{-5mm}
    \caption{Overview of MSGT-SL, consisting of four steps:
    {\it \textbf{1) Multi-view GNN}} learns local structural patterns from both SL data and multi-omics data,
    {\it\textbf{2) Cross-omics Sampling}} balances the scale of core and non-core genes in the SL and multi-omics data,
    {\it\textbf{3) Self-attention}} learns potential long-range gene dependencies,
    {\it\textbf{4) SL Prediction}} is implemented with pairwise gene features.
    }
	\label{Fig_1}
\vspace{-2mm}
\end{figure*}

More Recently, deep learning-based computational methods, especially relying on graph neural networks (GNNs) \cite{wan2020exp2sl,cai2020dual,long2021graph,hao2021prediction,zhang2021tumor,wang2021kg4sl}, have become highly popular in SL prediction.
GNNs \cite{kipf2016semi,velivckovic2018graph,zhao2022multi,zhao2022deep} use message passing to aggregate features in gene graphs, thus capturing non-linear relationships among genes.
\cite{cai2020dual} proposes a dual-dropout graph convolutional network (DDGCN), which is the first GNN-based model for SL prediction.
\cite{long2021graph} integrates various biological sources as extra features and applies a graph attention network (GAT) \cite{velivckovic2018graph} to further improve gene features.
\cite{wang2021kg4sl} and \cite{liu2022pilsl} introduce additional elements related to SL (e.g., biological processes, diseases, compounds, and omics data) to a knowledge graph (KG) and then complete SL prediction and explanation.
Although these GNN-based methods have yielded remarkable results, they still face two problems:
i) While GNN aggregates features beyond local neighbors by stacking layers, the exponentially growing receptive field with increasing depth may lead to performance issues, including over-smoothing \cite{oono2020graph} and over-squashing \cite{alon2020bottleneck}.
ii) Given the scale difference between large-scale multi-omics data and SL data, applying omics gene graphs with different non-SL gene relationships to improve SL prediction remains a non-trivial challenge.

To address these issues, we propose a multi-omics sampling-based graph transformer for SL prediction, namely MSGT-SL.
To avoid the limitations of GNNs, we use a shallow multi-view GNN to learn local structural information of gene relationships from both SL and multi-omics data.
Besides, inspired by graph transformers (GTs) \cite{dwivedi2020generalization,chen2022structure,rampavsek2022recipe} enabling communication among all nodes with just one layer, we concatenate features from all views for each gene and input them into a GT layer to capture potential long-range gene dependencies.
In particular, genes in the SL data are usually only a small subgroup of genes in each omics.
Thus, to maintain focus on core genes belonging to SL data in the self-attention of GT, we design a novel cross-omics sampling to moderately incorporate additional non-core genes.
Specifically, for each input, we start with a batch of core genes and run parallel random walk \cite{xia2019random} sampling for multiple omics gene graphs encompassing them.
After that, we get a balanced union of core and non-core genes in a structure-aware manner while also reducing the computational burden of self-attention.
Finally, gene features are concatenated pairwise to identify SL relationships, i.e., edge classification.

Our main contributions can be summarized as follows:

\begin{itemize}
\item The first attempt to capture long-range gene dependencies using the GT framework.
\item A new cross-omics gene sampling strategy for integrating SL and multi-omics information.
\item Extensive experiments on real-world datasets demonstrate the superiority of MSGT-SL, highlighting the advantages gained from GT and multi-omics data.
\end{itemize}
\section{Methodology}\label{Sec_2}
In this part, we detail the multi-omics sampling-based graph transformer for SL identification (MSGT-SL), whose overview is exhibited in Fig. \ref{Fig_1}.
All symbols are summarized in Table \ref{Tab_1}.
%
%
%
%

\subsection{Multi-view graph neural network}\label{Sec_2_1}
Typically, a SL dataset can be formulated as a graph network $G$$=$$\{V,E\}$, where $V$$=$$\{v_{1},...,v_{N}\}$ is a set containing $N$ gene nodes and $E$ is the set of edges formed by the SL relationships.
Since each omics from the multi-omics data can be formulated in a similar manner to $G$, the integration of SL with omics data can be expressed as a multi-view graph $\mathcal{G}$$=$$\{G_{1},...,G_{M}\}$ with $M$ views.
It is worth noting that $V$ and $E$ under different views are distinct, and only the first view contains SL edges $E_{1}\in G_{1}$ serving as ground-truth labels for model training and testing.

As a paradigm designed for analyzing graph-structured data, graph neural networks (GNNs) update the features of nodes by iteratively aggregating information along their edges, capturing the relationships and structural patterns among nodes. 
Besides, the scope of such aggregation is often broadened by deepening the layers.
To overcome the limitations arising from increasing depth while capturing local structures, we employ a  multi-view GNN (MVGNN) with fewer layers on the multi-view graph $\mathcal{G}$, which is a GNN expanded $M$ times along the view dimension.
Taking classical convolutional aggregation \cite{kipf2016semi} as an example, the feature updating of MVGNN at the $l$-th layer is as follows:
\begin{align}
	\mathbf{F}^{l}_{i}=\sigma(\mathbf{A}_{i}\mathbf{F}^{(l-1)}_{i}\mathbf{W}^{l}_{i}),
	\label{Eq_1}
\end{align}
where the superscript and subscript denote the indices for layer $l\in[1,L]$ and view $i\in[1,M]$.
$\mathbf{A}_{i}\in\mathbb{R}^{N\times N}$ is another abstract representation of $G_{i}$, which refers to the normalized adjacency matrix.
$\mathbf{F}^{l}_{i}\in\mathbb{R}^{N\times D_{l}}$ represents the output node feature matrix of the $l$-th layer, where $\mathbf{F}^{0}_{i}\in\mathbb{R}^{N\times D_{0}}$ corresponds to the initial gene features. 
$\mathbf{W}^{l}_{i}\in\mathbb{R}^{D_{(l-1)}\times D_{l}}$ denotes a view-layer-specific trainable feature transformation matrix.
Besides, $\sigma(\cdot)$ is a non-linear activation function, such as ReLU or Tanh.
After parallel aggregation in multiple views, MVGNN yields the final feature tensor $\mathcal{F}$$=$$\{\mathbf{F}^{L}_{1},...,\mathbf{F}^{L}_{M}\}$, where $L$ denotes the last layer index.

\begin{table}[t]
\caption{Definitions of all symbols.}
\label{Tab_1}
\resizebox{\columnwidth}{!}{
\begin{tabular}{r|l}
\hline
Symbol & Definition\\
\hline
$G$ & A graph network composed of gene data\\
$V$ & All gene nodes in a gene graph\\
$S$ & All gene nodes sampled from multi-omics data\\
$E$ & All edges formed by the gene relationships\\
\hline
$M$ & The total number of views of gene graphs\\
$N$ & The total number of samples in a gene set\\
$D$ & The dimension of a feature matrix or vector\\
$L$ & The total number of feature aggregation layers\\
\hline
$\mathbf{A}$ & The adjacency matrix of a gene graph\\
$\mathbf{F};\mathbf{E}$ & The gene feature representation matrices\\
$\mathbf{W}$ & The feature transformation or projection matrix\\
$\mathbf{P}$ & The probability transition matrix\\
$\mathbf{D}$ & The degree vector of gene nodes in a graph\\
$\mathbf{C}$ & The edge classification matrix for SL prediction\\
$\mathbf{Q};\mathbf{K};\mathbf{V}$ & The query, key, and value matrices in self-attention\\
\hline
$v$ & The node sample in a gene set or graph\\
$i;j;k;l$ & These lowercase letters represent different indices\\
\hline
$\sigma(\cdot)$ & The activation function, such as ReLU or Tanh\\
$\oplus(\cdot)$ & The concatenation operation\\
$\text{Attn}(\cdot)$ & The self-attention calculation\\
\hline
$\mathcal{G}$ & The multi-view graph from SL and omics data\\
$\mathcal{F}$ & The feature tensor of a multi-view gene graph\\
$\mathcal{E}$ & The feature tensor of pairwise concatenated genes\\
$\mathcal{Y}$ & The true label tensor of the predicted edges\\
$\mathcal{L}$ & The model training loss of MSGT-SL\\
\hline
\end{tabular}}
\end{table}

\subsection{Cross-omics sampling}\label{Sec_2_2}
Since the core gene set, $V_{1}$, required for SL prediction tasks, is usually a small subset of any omics gene set $V_{i}$\ ($i\in[2,M])$, the stacked MVGNN struggles to fuse non-core genes ($V_{i}\setminus V_{1}$) beyond the local scope of $V_{1}$ in $G_{i}$. 
In other words, the iterative hop-wise exploration based on $V_{1}$ tends to propagate excessive non-core
gene information to $V_{1}$, posing a threat to the primacy of core genes.
To elegantly incorporate auxiliary genes beyond SL data, we propose a cross-omics sampling strategy based on random walks \cite{xia2019random}.
Concretely, after aggregation, it starts with a batch of core genes and runs sampling across multiple omics views that contain them, as shown in Fig. \ref{Fig_1}.
For the $i$-th omics view, the transition probabilities  can be formulated as follows:
\begin{align}
	\mathbf{P}_{i}(v_{k}|v_{j})=\begin{cases}
	\frac{1}{\mathbf{D}_{i}(v_{j})}, &\mathbf{A}_{i}(v_{k},v_{j})>0\\
	0, &otherwise\end{cases},
	\label{Eq_2}
\end{align}
where $\mathbf{P}_{i}(v_{k}|v_{j})$ is an entry of the probability transition matrix $\mathbf{P}_{i}\in\mathbb{R}^{N\times N}$, symbolizing the probability of transitioning from the current gene node $v_{j}$ to the next one $v_{k}$.
$\mathbf{D}_{i}\in\mathbb{R}^{1\times N}$ stands for the degree vector of gene nodes in $V_{i}$, where $\mathbf{D}_{i}(v_{j})$ is the number of edges originated from
 $v_{j}$.
Through serial transitions, we traverse non-SL relationships to uncover a set of genes that have potential associations with the core genes, denoted as $S_{i}$.
After parallel walks in diverse omics views, we treat the union $S$$=$$\cup_{i=2}^{M}S_{i}$ as the output genes.
In this way, we strike a balance between the scale of core and non-core genes, especially when dealing with large-scale multi-omics data.

\subsection{Self-attention}\label{Sec_2_3}
After obtaining the feature tensor $\mathcal{F}$ and the sampled cluster $S$ of gene nodes, we concatenate the features of all views based on the gene indices and filter out irrelevant gene samples using $S$, which is formulated as follows:
\begin{align}
	\mathbf{\overline{F}}(v_{j})=\mathbf{F}_{1}^{L}(v_{j})\oplus\mathbf{F}_{2}^{L}(v_{j})\cdots\oplus\mathbf{F}_{M}^{L}(v_{j}),\ s.t.\ v_{j}\in S,
	\label{Eq_3}
\end{align}
where $\mathbf{F}_{i}^{L}(v_{j})\in\mathbb{R}^{1\times D_{L}}$ is the $D_{L}$-dimensional vector of gene node $v_{j}$ in the $i$-th view, and $\oplus$ is an end-to-end splicing of two vectors.
Thus, $\mathbf{\overline{F}} \in\mathbb{R}^{N'\times D'}$ is the integrated feature matrix for $S$, where $N'$ represents the total number of sampled genes and $D'$$=$$MD_{L}$ represents the node feature dimension.

Nonetheless, $\mathbf{\overline{F}}$ solely equips the local structural information from multiple views, disregarding the long-range dependencies in the gene set $S$.
Drawing inspiration from graph transformers (GTs) \cite{dwivedi2020generalization}, which allow input nodes to communicate with each other without hop restrictions, here we utilize a single standard self-attention layer to obtain implicit global structural patterns. 
We project the input $\mathbf{\overline{F}}$ into three components of self-attention:
\begin{align}
	\mathbf{Q}=\mathbf{\overline{F}}\mathbf{W}_{\alpha},\ \mathbf{K}=\mathbf{\overline{F}}\mathbf{W}_{\beta},\ \mathbf{V}=\mathbf{\overline{F}}\mathbf{W}_{\gamma},
	\label{Eq_4}
\end{align}
where $\mathbf{W}_{\alpha}$, $\mathbf{W}_{\beta}$, and $\mathbf{W}_{\gamma}$ are all $D'\times D''$ projection matrices. $\mathbf{Q}$, $\mathbf{K}$, and $\mathbf{V}$ denote query, key, and value matrices of $N'\times D''$ shape. 
After that, the calculation of self-attention is as follows:
\begin{align} 
    \text{Attn}(\mathbf{\overline{F}})=\text{softmax}(\mathbf{\overline{A}})\mathbf{V},\ s.t.\ \mathbf{\overline{A}}=\frac{\mathbf{Q}\mathbf{K}^{\top}}{\sqrt{D''}},
	\label{Eq_5}
\end{align}
where $\mathbf{\overline{A}}\in\mathbb{R}^{N'\times N'}$ denotes a matrix with similarities between queries and keys, and $\text{softmax}(\mathbf{\overline{A}})$ stores attention weights that determine the importance or relevance of each input gene node with respect to the current node.
For brevity, we omit the direct extension of the single-head calculation to its multi-head form, which would yield more stable results in practical experiments.
Moreover, we cleverly dealt with two possible problems of the GT: i) Since the multi-omics matrices $\{\mathbf{F}^{L}_{i}|i\in[2,M]\}$ spliced after $\mathbf{F}^{L}_{1}$ can essentially be regarded as a new type of structural encoding (SE) \cite{ying2021transformers,chen2022structure}, self-attention effectively understands permutation-invariant gene nodes while retaining gene identity awareness for SL prediction.
ii) Cross-omics sampling not only ensures the dominance of core genes, but also reduces the input scale $N'$ of self-attention and avoids inefficient GT calculation.

\begin{algorithm}[t]
\caption{Multi-omics Sampling-based Graph Transformer for synthetic lethality (MSGT-SL)}
\label{Alg_1}
{\bf Input:} A multi-view gene graph $\mathcal{G}$ and true SL edge labels $\mathcal{Y}$\\
{\bf Output:} A MSGT-SL proficient in predicting SL relationships
    \begin{algorithmic}[1]
    \State $\triangleright$ \textbf{Step 1: MVGNN} $\gets$ Sec. \ref{Sec_2_1}
    \State $\{\mathbf{F}_{i}^{l}|i\in[1,M],\ l\in[0,L]\} \gets$ Eq.~(\ref{Eq_1}) // {\it{local aggregation}}
    \For{$batch\_index$$=$$1$ to $last\_batch$}
        \State $\triangleright$ \textbf{Step 2: Cross-omics sampling} $\gets$ Sec. \ref{Sec_2_2}
        \State $S\gets\{S_{i}\gets\mathbf{P}_{i}|i\in[2,M]\}\gets$ Eq.~(\ref{Eq_2}) // {\it{random walk}}
        \State $\triangleright$ \textbf{Step 3: Self-attention} $\gets$ Sec. \ref{Sec_2_3}
        \State $\mathbf{\overline{F}}\gets$ Eq.~(\ref{Eq_3}) // {\it{concatenation of features from all views}}
        \State $\mathbf{Q},\mathbf{K},\mathbf{V}\gets$ Eq.~(\ref{Eq_4}) // {\it{key components of self-attention}}
        \State $\text{Attn}(\mathbf{\overline{F}})\gets$ Eq.~(\ref{Eq_5}) // {\it{capture long-range dependencies}}
        \State $\triangleright$ \textbf{Step 4: SL prediction} $\gets$ Sec. \ref{Sec_2_4}
        \State $\mathcal{E},\mathbf{E}\gets$ Eqs.~(\ref{Eq_6}, \ref{Eq_7}) // {\it{pairwise splicing of gene features}}
        \State $\mathcal{L}\gets$ Eq.~(\ref{Eq_8}) // {\it{cross-entropy loss of edge classification}}
    \EndFor
\end{algorithmic}
\end{algorithm}

\subsection{SL prediction}\label{Sec_2_4}
To realize SL prediction, we first extract relevant genes from the feature matrix $\text{Attn}(\mathbf{\overline{F}})\in\mathbb{R}^{N'\times D''}$ output by the GT layer:
\begin{align}
	\mathbf{E}(v_{j})=\text{Attn}(\mathbf{\overline{F}})(v_{j}),\ s.t.\ v_{j}\in V_{1}\cap S,
	\label{Eq_6}
\end{align}
where $\mathbf{E}(v_{j})\in\mathbb{R}^{1\times D''}$ indicates the feature vector of the gene $v_{j}$ belonging to the core set $V_{1}$. 
Then, we concatenate the gene features in a pairwise manner to yield the edge features of $G_{1}$:  
\begin{align}
	\mathcal{E}(v_{j},v_{k})=\mathbf{E}(v_{j})\oplus\mathbf{E}(v_{k}),\ s.t.\ v_{j},v_{k}\in V_{1}\cap S,
	\label{Eq_7}
\end{align}
where $\mathcal{E}(v_{j},v_{k})$ indicates the feature vector of the virtual edge between the core genes $v_{j}$ and $v_{k}$, whose vector dimension is $2D''$.
Subsequently, we perform the binary classification of SL relationships and use a cross-entropy loss function to optimize:
\begin{align}
    \footnotesize
	\mathcal{L}=-\frac{1}{(N'')^{2}}\sum_{j=1}^{N''}\sum_{k=1}^{N''}\text{log}(\text{softmax}(\mathcal{E}(v_{j},v_{k})\mathbf{C}))\mathcal{Y}(v_{j},v_{k}),
	\label{Eq_8}
\end{align}
where $\mathbf{C}\in\mathbb{R}^{2D''\times 2}$ can be regarded as an edge classifier, and $\mathcal{Y}(v_{j},v_{k})\in\mathbb{R}^{2\times 1}$ is the transposed label vector corresponding to $\mathcal{E}(v_{j},v_{k})$.
In addition, $N''$$=$$|V_{1}\cap S|$ denotes the total number of core genes in the sampled cluster.
By continuously adjusting the model parameters via backpropagation, the construction of MSGT-SL is eventually completed.
The algorithm overview is summarized in Alg. \ref{Alg_1}.
\section{Experiments}\label{Sec_3}
In this part, we conduct evaluation experiments to verify the performance of MSGT-SL.
%
%
%
%
%
Three aspects about MSGT-SL can be condensed as follows:
\begin{itemize}
\item It achieves the best results across two SL prediction tasks.
\item Each of its constituents contributes to result improvement.
\item It is insensitive to key hyperparameters and very efficient.
\end{itemize}

\subsection{Datasets and metrics}\label{Sec_3_1}
The experiments involve the following two real datasets for SL prediction, with statistical information detailed in Table \ref{Tab_2}.
Specifically, following an efficient processing strategy \cite{fan2023multi}, we utilize the mapping system proposed by \cite{horlbeck2018mapping} to collect cancer cell-specific SL data.
Using a double-knockdown CRISPR (the acronym for clustered regularly inter-spaced short palindromic repeats) interference technology, we quantify gene interactions of 448 and 387 specified core genes from two cell lines ($\mathbf{K562}$ \& $\mathbf{Jurkat}$) respectively, and then identify SL gene pairs based on interaction scores below -3.
Taking the K562 as an example, we build the SL view in the dataset named after it, whose graph topology will provide valuable information about unknown SL relationships within this specific cell line.
Besides, we consider additional omics views from general population analysis rather than specific to one cell line.
Concretely, here we integrate four categories of omics views from different dataset sources:
i) We yield two omics views via the biological general repository for interaction datasets (BioGRID) \cite{oughtred2019biogrid}, which reveal the {\it{Physical}} and {\it{Genetic}} interactions between genes from multiple different cell lines.
ii) We calculate the Pearson correlation among genes from the cancer cell line encyclopedia (CCLE) \cite{ghandi2019next} {\it{Expression}} profiles and connect significantly correlated gene pairs $<0.01$.
iii) We construct a co-{\it{Essentiality}} view with the CRISPR data from the cancer dependency map portal (DepMap) \cite{meyers2017computational}, which reflects common patterns of interactions among genes.
Overall, these cell-independent views also provide valuable information for predicting SL pairs specific to the K562 or Jurkat cell lines. 
For any gene in any view, its characteristics include expression, copy number, mutation, and essentiality.

To quantify the performance of SL prediction, we apply the following evaluation metrics, namely accuracy (Acc), F1-score (F1), and area under the receiver operating characteristic curve (ROC-AUC).

\renewcommand{\arraystretch}{1.25}
\begin{table}[t]
\caption{
Statistics of datasets. Only SL views have SL edges as ground-truth labels for training and testing of SL prediction.}
\label{Tab_2}
\resizebox{\columnwidth}{!}{%
\begin{tabular}{c|cccccc}
\hline
\multirow{2}{*}{Data} & \multirow{2}{*}{Number} & \multirow{2}{*}{SL view} & \multicolumn{4}{c}{Multi-omics views} \\ \cline{4-7} 
 &  &  & Phy & Gen & Exp & Ess \\ \hline
\multirow{2}{*}{K562} & Node & 448 & 19094 & 4613 & 12644 & 14347 \\
 & Edge & 1523 & 1411290 & 21828 & 1168026 & 445540 \\ \hline
\multirow{2}{*}{Jurkat} & Node & 387 & 19094 & 4613 & 12644 & 14347 \\
 & Edge & 373 & 1411290 & 21828 & 1168026 & 445540 \\ \hline
\end{tabular}%
}
\end{table}
\renewcommand{\arraystretch}{1}

\begin{table*}[t]
\caption{
Experimental results on two SL prediction datasets under the leave-gene-pair-out evaluation setting, where the best and second-best results are shown in \textbf{bold} and {\it{italics}}.
The last column and row represent the mean of all metrics and the gain (\%$\uparrow$) of the best result compared to the second-best result.
}
\label{Tab_3}
\resizebox{\textwidth}{!}{%
\begin{tabular}{c|ccc|ccc|c}
\hline
\multirow{2}{*}{Model} & \multicolumn{3}{c|}{K562} & \multicolumn{3}{c|}{Jurkat} & \multirow{2}{*}{Mean} \\ \cline{2-7}
 & Acc & F1 & ROC-AUC & Acc & F1 & ROC-AUC &  \\ \hline
GCN & 0.696 ± 0.003 & 0.694 ± 0.004 & 0.790 ± 0.003 & 0.628 ± 0.003 & 0.621 ± 0.005 & 0.711 ± 0.006 & 0.690 \\
GAT & 0.704 ± 0.004 & 0.700 ± 0.004 & 0.796 ± 0.005 & 0.631 ± 0.006 & 0.627 ± 0.005 & 0.721 ± 0.010 & 0.696 \\
DDGCN & 0.702 ± 0.006 & 0.701 ± 0.005 & 0.793 ± 0.005 & 0.628 ± 0.003 & 0.624 ± 0.002 & 0.717 ± 0.016 & 0.694 \\
GCATSL & 0.727 ± 0.002 & 0.722 ± 0.002 & 0.801 ± 0.004 & 0.647 ± 0.006 & 0.642 ± 0.007 & 0.734 ± 0.013 & 0.712 \\
KG4SL & 0.711 ± 0.003 & 0.710 ± 0.004 & 0.798 ± 0.003 & 0.636 ± 0.003 & 0.634 ± 0.002 & 0.724 ± 0.019 & 0.702 \\
PiLSL & 0.730 ± 0.002 & 0.726 ± 0.004 & 0.805 ± 0.010 & 0.674 ± 0.003 & 0.672 ± 0.002 & 0.742 ± 0.021 & 0.724 \\
MVGCN-iSL & \textit{0.734 ± 0.002} & \textit{0.730 ± 0.002} & \textit{0.808 ± 0.007} & \textit{0.677 ± 0.003} & \textit{0.675 ± 0.004} & \textit{0.749 ± 0.021} & \textit{0.728} \\ \hline
MSGT-SL & \textbf{0.766 ± 0.005} & \textbf{0.764 ± 0.007} & \textbf{0.835 ± 0.012} & \textbf{0.737 ± 0.006} & \textbf{0.733 ± 0.006} & \textbf{0.803 ± 0.008} & \textbf{0.773} \\
Gain (\%) & 3.3$\uparrow$ & 3.4$\uparrow$ & 2.7$\uparrow$ & 6.0$\uparrow$ & 5.8$\uparrow$ & 5.4$\uparrow$ & 4.5$\uparrow$ \\ \hline
\end{tabular}%
}
\end{table*}

\begin{table*}[t]
\caption{Experimental results on two SL prediction datasets under the leave-gene-out evaluation setting.}
\label{Tab_4}
\resizebox{\textwidth}{!}{%
\begin{tabular}{c|ccc|ccc|c}
\hline
\multirow{2}{*}{Model} & \multicolumn{3}{c|}{K562} & \multicolumn{3}{c|}{Jurkat} & \multirow{2}{*}{Mean} \\ \cline{2-7}
 & Acc & F1 & ROC-AUC & Acc & F1 & ROC-AUC &  \\ \hline
GCN & - & - & - & - & - & - & - \\
GAT & - & - & - & - & - & - & - \\
DDGCN & - & - & - & - & - & - & - \\
GCATSL & 0.515 ± 0.006 & 0.511 ± 0.005 & 0.537 ± 0.020 & 0.508 ± 0.012 & 0.507 ± 0.012 & 0.512 ± 0.014 & 0.515 \\
KG4SL & 0.510 ± 0.003 & 0.505 ± 0.003 & 0.528 ± 0.003 & 0.508 ± 0.012 & 0.504 ± 0.011 & 0.507 ± 0.006 & 0.510 \\
PiLSL & 0.539 ± 0.012 & 0.530 ± 0.011 & 0.545 ± 0.014 & 0.526 ± 0.021 & \textit{0.518 ± 0.019} & \textit{0.541 ± 0.012} & 0.533 \\
MVGCN-iSL & \textit{0.541 ± 0.003} & \textit{0.535 ± 0.003} & \textit{0.554 ± 0.005} & \textit{0.526 ± 0.021} & 0.515 ± 0.016 & 0.533 ± 0.013 & \textit{0.534} \\ \hline
MSGT-SL & \textbf{0.567 ± 0.003} & \textbf{0.560 ± 0.002} & \textbf{0.572 ± 0.004} & \textbf{0.570 ± 0.012} & \textbf{0.557 ± 0.009} & \textbf{0.609 ± 0.005} & \textbf{0.572} \\
Gain (\%) & 2.6$\uparrow$ & 2.5$\uparrow$ & 1.8$\uparrow$ & 4.4$\uparrow$ & 3.9$\uparrow$ & 6.8$\uparrow$ & 3.8$\uparrow$ \\ \hline
\end{tabular}%
}
\vspace{-1mm}
\end{table*}

\subsection{Baselines}\label{Sec_3_2}
We compare MSGT-SL with several state-of-the-art (SOTA) GNN-based baselines for SL prediction: $\textbf{GCN}$ and $\textbf{GAT}$ \cite{kipf2016semi,velivckovic2018graph} are designed for feature learning on homogeneous graphs. 
$\textbf{DDGCN}$ \cite{cai2020dual} designs a new dual-dropout mechanism to tackle overfitting but lacks external sources of information.
$\textbf{GCATSL}$ \cite{long2021graph} introduces biological data, including biological processes, cellular components, and protein-protein interactions as inputs, and applies a dual-attention mechanism to obtain gene features from different input graphs.
$\textbf{KG4SL}$ and $\textbf{PiLSL}$ \cite{wang2021kg4sl,liu2022pilsl} use knowledge graphs (KGs) as model inputs.
KG4SL employs the attention mechanism to calculate the weights of different types of nodes and edges in each aggregation layer.
PiLSL constructs locally closed sub-graphs for each pair of genes and integrates multi-omics data to acquire more expressive features for more robust SL prediction.
$\textbf{MVGCN-iSL}$ \cite{fan2023multi} employs a multi-view GCN to integrate gene features from multi-omics data to serve SL prediction.
All models predict lethality only for core genes.

\subsection{Settings}\label{Sec_3_3}
MSGT-SL\footnote{All data and codes are available at https://github.com/MSGT-SL/MSGT-SL} implements a shallow MVGNN with $L$=2 layers based on GCN, with dimensions $D_{1}$$=$$128$ and $D_{2}$$=$$64$.
For any batch, it starts with 100 core genes $\in V_{1}$ and then runs random walks with the length of 10 in four omics views $V_{i}\ (i\in[2,5])$.
After obtaining a sampling set $S$, it performs resampling while limiting the total number of genes to $N'$$=$$500$.
Next, it sets the projection dimension and the number of heads in self-attention to 64 and 4.
Since SL edges are exclusive to the SL view, the samples from the SL view are divided into training, validation, and test sets in a ratio of 7:1:2.
It takes the Adam optimizer to optimize parameters with a learning rate of 0.0001.
To address overfitting, it adopts an early stopping technique.
All codes are based on Python and repeated 3 times on the NVIDIA GeForce RTX 3080 (10240 MiB) GPU.

We employ two evaluation settings: leave-gene-pair-out and leave-gene-out.
The former divides the training, validation, and test sets by edges, exposing the genes involved in all test edges to training (transductive).
The latter divides the three sets based on genes, with no exposure of test genes in training (inductive).

\begin{table*}[t]
\caption{Ablation results of the three modules in MSGT-SL and multi-omics data under the leave-gene-pair-out setting.
The best and worst results are indicated as \textbf{bold} and \underline{underlined}, where ``w/o'' denotes ``without''.
The last column and row illustrate the mean of all metrics and the loss (\%$\downarrow$) of the best result compared to the worst result.
}
\label{Tab_5}
\resizebox{\textwidth}{!}{%
\begin{tabular}{c|ccc|ccc|c}
\hline
\multirow{2}{*}{Model} & \multicolumn{3}{c|}{K562} & \multicolumn{3}{c|}{Jurkat} & \multirow{2}{*}{Mean} \\ \cline{2-7}
 & Acc & F1 & ROC-AUC & Acc & F1 & ROC-AUC &  \\ \hline
w/o MVGNN & {\ul 0.605 ± 0.012} & {\ul 0.598 ± 0.011} & {\ul 0.637 ± 0.007} & {\ul 0.568 ± 0.027} & {\ul 0.541 ± 0.032} & {\ul 0.586 ± 0.034} & {\ul 0.589} \\
w/o Sampling & 0.765 ± 0.008 & 0.759 ± 0.009 & 0.824 ± 0.012 & 0.726 ± 0.003 & 0.724 ± 0.003 & 0.802 ± 0.009 & 0.766 \\
w/o GT & 0.727 ± 0.002 & 0.723 ± 0.001 & 0.803 ± 0.007 & 0.647 ± 0.006 & 0.643 ± 0.007 & 0.763 ± 0.014 & 0.717 \\
w/o Omics & 0.709 ± 0.006 & 0.706 ± 0.006 & 0.797 ± 0.008 & 0.672 ± 0.006 & 0.667 ± 0.006 & 0.769 ± 0.012 & 0.720 \\ \hline
MSGT-SL & \textbf{0.766 ± 0.005} & \textbf{0.764 ± 0.007} & \textbf{0.835 ± 0.012} & \textbf{0.737 ± 0.006} & \textbf{0.733 ± 0.006} & \textbf{0.803 ± 0.008} & \textbf{0.773} \\
Loss (\%) & 16.1$\downarrow$ & 16.6$\downarrow$ & 19.8$\downarrow$ & 16.9$\downarrow$ & 19.2$\downarrow$ & 21.7$\downarrow$ & 18.4$\downarrow$ \\ \hline
\end{tabular}%
}
\end{table*}

\begin{table*}[t]
\caption{Ablation results of the three modules in MSGT-SL and multi-omics data under the leave-gene-out setting.}
\label{Tab_6}
\resizebox{\textwidth}{!}{%
\begin{tabular}{c|ccc|ccc|c}
\hline
\multirow{2}{*}{Model} & \multicolumn{3}{c|}{K562} & \multicolumn{3}{c|}{Jurkat} & \multirow{2}{*}{Mean} \\ \cline{2-7}
 & Acc & F1 & ROC-AUC & Acc & F1 & ROC-AUC &  \\ \hline
w/o MVGNN & {\ul 0.526 ± 0.003} & {\ul 0.516 ± 0.006} & {\ul 0.533 ± 0.007} & 0.535 ± 0.012 & 0.522 ± 0.013 & 0.554 ± 0.018 & {\ul 0.531} \\
w/o Sampling & 0.562 ± 0.006 & 0.556 ± 0.006 & 0.569 ± 0.007 & 0.561 ± 0.032 & 0.552 ± 0.029 & 0.580 ± 0.017 & 0.563 \\
w/o GT & 0.536 ± 0.003 & 0.531 ± 0.004 & 0.548 ± 0.010 & {\ul 0.526 ± 0.021} & {\ul 0.514 ± 0.012} & {\ul 0.534 ± 0.009} & 0.531 \\
w/o Omics & - & - & - & - & - & - & - \\ \hline
MSGT-SL & \textbf{0.567 ± 0.003} & \textbf{0.560 ± 0.002} & \textbf{0.572 ± 0.004} & \textbf{0.570 ± 0.012} & \textbf{0.557 ± 0.009} & \textbf{0.609 ± 0.005} & \textbf{0.572} \\
Loss (\%) & 4.1$\downarrow$ & 4.4$\downarrow$ & 3.7$\downarrow$ & 4.4$\downarrow$ & 4.3$\downarrow$ & 7.5$\downarrow$ & 4.3$\downarrow$ \\ \hline
\end{tabular}%
}
\end{table*}

\subsection{Prediction results}\label{Sec_3_4}
\vspace{-1mm}
We compare the performance of our MSGT-SL with various existing SOTA baselines for SL prediction on two cell-specific gene datasets that incorporate multi-omics data.
We present the edge classification results under the leave-gene-pair-out setting in Table \ref{Tab_3}, from which we sum up the following observations:
1) The GCN model performs worst in both cell lines, implying that local feature aggregation on SL data alone is inapplicable.
2) GAT and DDGCN surpass GCN because they either employ better attention-based feature aggregation or discard nodes and edges to regulate aggregation objects.
Such observations imply that the wise use of feature aggregation benefits SL prediction.
3) GCN, GAT, and DDGCN perform worse than baselines that introduce additional information, indicating the significance of auxiliary data in SL prediction.
4) KG-based KG4SL is weaker than GCATSL with added biological data. 
Apart from inherent differences between model architectures, we argue that another possible explanation is that the complicated information in KG is not completely suitable for SL analysis.
5) PiLSL, which is also based on KG, outperforms GCATSL. 
It is not inconsistent with the previous conclusion because PiLSL additionally uses multi-omics data to enhance the gene features.
Combined with the superb performance of MVGCN-iSL, we can conclude that introducing informative multi-omics data can indeed reinforce SL prediction.
6) Our MSGT-SL achieves SOTA results across all metrics, with a mean 4\% improvement over the second-best result.
This finding is as expected.
On the one hand, in addition to 2-hop local aggregation, we introduce the graph transformer (GT) to capture long-range dependencies among genes, which is ignored by other models.
On the other hand, we moderately sample non-core genes across several large-scale omics views,
which harmoniously incorporate structure-aware auxiliary data while ensuring the dominance of core genes.

Under the leave-gene-pair-out setting, the endpoint genes of all test edges are already present in training, akin to identifying some new SL edges in one whole fixed gene graph with known partial SL gene pairs.
In contrast, with a leave-gene-out setting, the test genes are completely unseen to the trained model, thus serving to evaluate the capacity of the models to generalize to SL prediction of novel genes.
The corresponding experimental results under leave-gene-out are shown in Table \ref{Tab_4}, displaying similar patterns as the leave-gene-pair-out setting.
Notably, the transductive baselines, GCN, GAT, and DDGCN, which do not incorporate additional information are unable to perform novel edge prediction.
GCATSL and KG4SL show poor results on all metrics, indicating weak generalizability, while the other three models, PiLSL, MVGCN-iSL, and MSGT-SL, which integrate the same multi-omics data, all have better generalization.
Such observations again justify that additional auxiliary information for SL prediction can facilitate more general gene embeddings, especially for omics data directly related to gene studies.
Next, another noteworthy observation is that MSGT-SL demonstrates remarkable breakthroughs, beating all baseline models.
This is because these previous GNN-based SL methods typically learn from static gene graph networks constrained by predetermined structural induction biases.
For the GT layer, every batch of its inputs and aggregations are randomly and dynamically formed, so it constantly forces MSGT-SL to learn how to deal with new genes during training.
Hence, the best generality of MSGT-SL makes it an ideal auxiliary tool to accelerate real SL prediction applications in industries where the gene pairs to be identified are often unfamiliar.
Overall, the proposed MSGT-SL achieves SOTA performance under both evaluation settings.

\subsection{Ablation analysis}\label{Sec_3_5}
\vspace{-1mm}
To further investigate the role of each module in MSGT-SL, we perform ablation analysis and then showcase the results in Table \ref{Tab_5} and Table \ref{Tab_6}.
We acquire four observations as follows: 
1) Removing any module hurts the performance of MSGT-SL, affirming their validity.
2) The absence of the MVGNN module leads to the most significant drop in average results.
The reason is that MVGNN plays a key role in reinforcing the initial gene features, even if it brings strict structural inductive biases \cite{chen2022structure}.
3) Less performance impact after replacing random walks with random sampling in cross-omics sampling.
This is because the random walk sampling has strong randomness while also being structure-aware, then each sampled gene set implicitly encodes non-SL gene relationships within its omics view.
4) Removing the GT module under the leave-gene-out setting also yields the largest performance drop, which is consistent with the analysis of Table \ref{Tab_4}.
As previous studies in \cite{chen2022structure,hussain2022global} have pointed out, GT dynamically learns the long- and short-range dependencies among input nodes, relaxing the inductive biases constraints of the input gene graphs. 
Moreover, due to the implementation of cross-omics sampling, our GT deals with irregular gene nodes, which alleviates the influence of inductive biases and improves the model generalizability.
In addition to the internal modules, we also fulfill an ablation analysis on the multi-omics data and obtain two observations:
5) Upon removing the omics data, the results under the leave-gene-pair-out evaluation setting surpass the results of using GCN alone.
This is because after removing omics data, our MSGT-SL degenerates into a single-view GCN followed by GT in the SL view, thereby capturing longer-range gene dependencies and higher-quality gene features than GCN.
6) The model without multi-omics data cannot realize effective training and SL prediction under the leave-gene-out evaluation setting, which once again emphasizes the significance of omics data for enhancing generalizability.

We summarize the driving inspirations of all components in MSGT-SL as follows:
To circumvent restricted generalizability using SL data alone, we leverage informative multi-omics data.
To learn from the multi-view gene graph network, we leverage MVGNN, the de-facto multi-view graph learning paradigm, to discover local structural patterns of various gene relationships.
To discover the global structural patterns of gene relationships and avoid issues (e.g. over-smoothing) that GNN is susceptible to, we apply a GT that satisfies any spacing between two genes for communication.
To prevent deviation from cell-specific SL predictions by GT embedding too many non-core gene features into core genes due to the scale imbalance between omics data and SL data, we design a new random walk-based cross-omics sampling strategy.
Serving as the bridge between MVGNN and GT, this sampling reduces the calculation burden of each batch of self-attention and improves the model generalizability.
Also, because we concatenate the outputs from all views of MVGNN and its feature assembly of omics views is a natural structural encoding (SE), there is no need to introduce additional position encoding in GT.
Overall, the introduction of all parts in MSGT-SL is feasible and effective.

\begin{figure}[t]
\centering
\subfigure{
    \begin{minipage}[t]{0.479\columnwidth}
        \includegraphics[width=\textwidth]{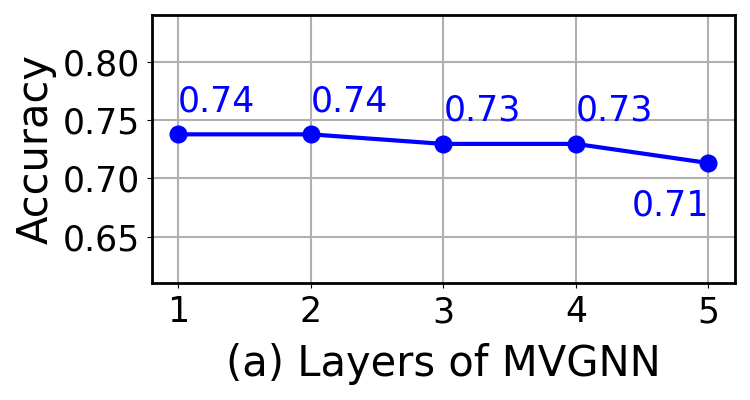} \\
        \includegraphics[width=\textwidth]{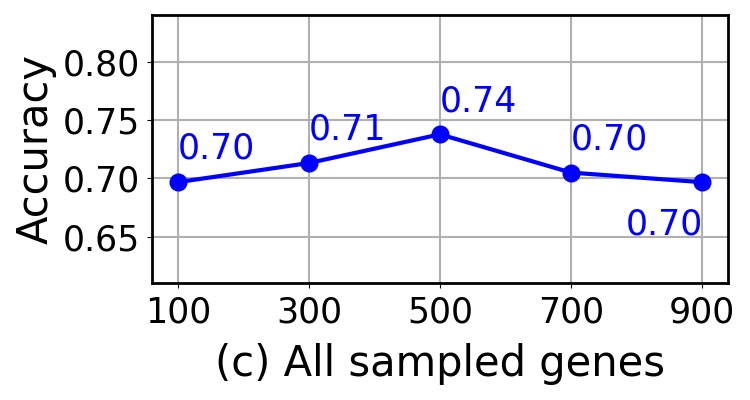}
    \end{minipage}}
\subfigure{
    \begin{minipage}[t]{0.479\columnwidth}
        \includegraphics[width=\textwidth]{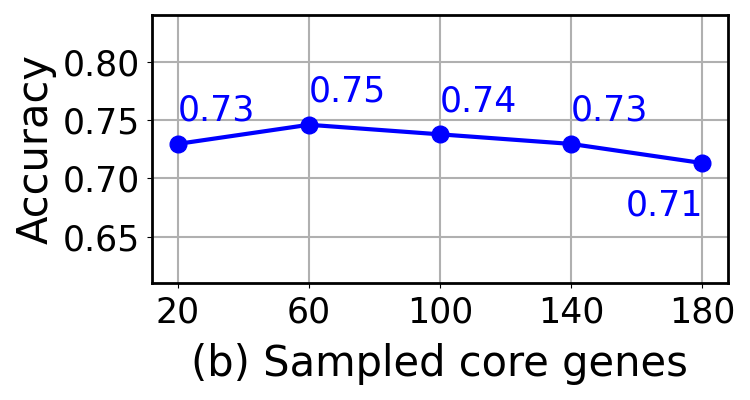} \\
        \includegraphics[width=\textwidth]{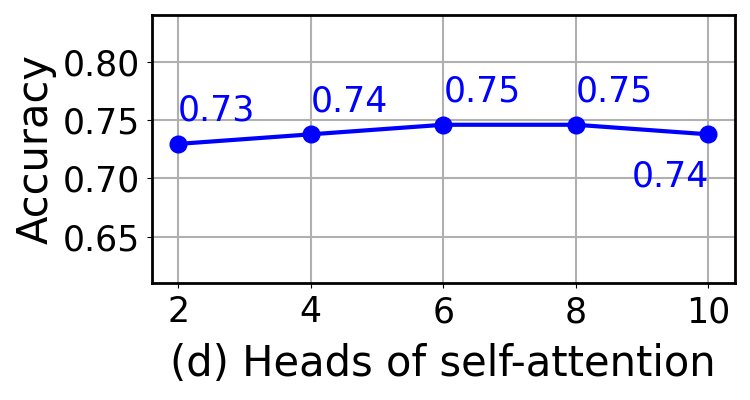}
    \end{minipage}}
\vspace{-2.5mm}
\caption{
Results of sensitivity analysis for four hyperparameters.
}
\label{Fig_2}
\end{figure}

\subsection{Hyperparameter analysis}\label{Sec_3_6}
To assess the impact of crucial parameters in MSGT-SL on performance, we realize hyperparameter sensitivity analysis on Jurkat in the leave-gene-pair-out setting.
The four hyperparameters include the number of MVGNN layers, the count of core genes and the total number of genes in cross-omics sampling, and the number of self-attention heads in GT.
We exhibit their results in Fig. \ref{Fig_2}, and draw four observations:
1) As the number of MVGNN layers increases, the accuracy depicts a downward trend, indicating that widening the aggregation range hop-by-hop is suboptimal.
2) With an increasing number of core genes, the model performance first increases and then decreases.
This change is attributed to the higher number of core genes leading to a reduction in non-core genes.
Therefore, a lower proportion of non-core genes adversely affects the model performance.
3) The total number of sampled gene nodes also triggers a similar hump-shaped change.
The consistent number of core genes and increasing proportion of non-core genes affect performance by disrupting the dominance of cell-specific core genes.
At a total number of 100, all genes are from SL data, equivalent to using MSGT-SL without multi-omics data.
4) As the number of self-attention heads increases, the model learns features from more perspectives and subsequently integrates them, thereby leading to performance improvement. 
However, after attaching enough heads, increasing built-in parameters without more genes risks overfitting and does not improve performance.
Overall, the key hyperparameters from MSGT-SL cause modest and reasonable performance changes.

\begin{figure}[t]
\centering
\includegraphics[width=\columnwidth]{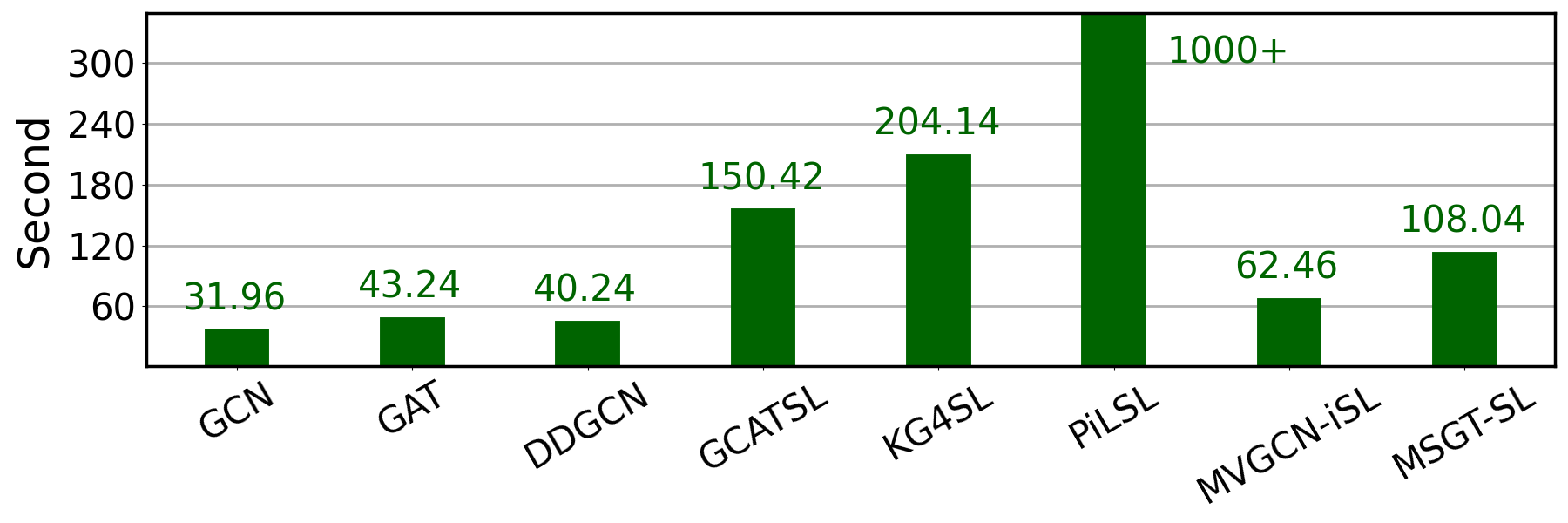}
\vspace{-4.5mm}
\caption{
Time required for all models to run on the Jurkat dataset under the leave-gene-pair-out setting.
}
\label{Fig_3}
\end{figure}

\subsection{Computational complexity analysis}\label{Sec_3_7}
Since the efficiency of dry-lab strategies is also an essential factor in real SL prediction, here we analyze the computational complexity of MSGT-SL, which is mainly affected by the main parts described in Sec. \ref{Sec_2}.
Firstly, the multi-view convolutional aggregation on SL data as well as multi-omics data incurs the cost of $\mathcal{O}(MLND^{2})$.
Secondly, sampling operations in cross-omics sampling cost $\mathcal{O}(M|S_{i}|)$, where $|S_{i}|$ indicates the length of random walk in an omics view.
Thirdly, operations such as feature concatenation and self-attention calculation incur a cost of $\mathcal{O}(HN^{2}D)$, where $H$ is the number of self-attention heads.
Finally, SL prediction costs $\mathcal{O}(2N)$.
Therefore, the complexity of MSGT-SL mainly arises from GNNs, similar to other GNN-based strategies.
For a more intuitive comparison, Fig. 3 shows the runtime of our MSGT-SL and all baselines on Jurkat under the leave-gene-pair-out setting.
Even without the fastest speed, MSGT-SL exhibits competitive efficiency.
Given the promising performance gains, investing in reasonable time resources for MSGT-SL is justified.
\section{Conclusion}\label{Sec_4}
In this paper, we propose a new multi-omics sampling-based graph transformer for SL prediction (MSGT-SL) of gene pairs.
Unlike existing computational methods that solely apply GNN for gene analysis, MSGT-SL adopts the GT framework to learn potential long-range dependencies among genes.
Then, MSGT-SL introduces additional auxiliary genes in a moderate manner through cross-omics sampling, ensuring the dominance of core genes used for SL prediction.
Such a random sampling strategy also promotes GT to alleviate inductive bias and computational efficiency issues.
Comprehensive experiments demonstrate that MSGT-SL substantially improves SL prediction with excellent generalizability and efficiency.
In future work, we will harness MSGT-SL as an auxiliary tool to expedite SL identification and combine it with reliable web-lab verification to achieve more practical deployment and application.
\section*{Acknowledgment}\label{Sec_5}
This research is supported by the Strategic Priority Research Program of Chinese Academy of Sciences (CAS) under Grant XDC02040400, National Natural Science Foundation of China under Grant 62002007, Natural Science Foundation of Beijing Municipality through Grant 4222030.

\end{document}